\documentclass[10pt, a4paper]{article}

\usepackage[]{lrec-coling2024} 

\usepackage{times}
\usepackage{latexsym}

\usepackage[T1]{fontenc}

\usepackage[utf8]{inputenc}

\usepackage{inconsolata}

\usepackage{graphicx}
\usepackage{amsmath}
\usepackage{amsfonts}
\usepackage{amssymb}
\usepackage{array}
\usepackage{amsthm}
\usepackage[ruled,vlined]{algorithm2e}
\usepackage{booktabs}
\usepackage{multirow}
\usepackage{multicol}
\usepackage{times}
\usepackage{latexsym}
\usepackage{color, soul}
\usepackage{graphicx}
\usepackage{amsmath,amssymb}
\usepackage{diagbox}

\usepackage{microtype}
\usepackage{booktabs}
\usepackage{multirow}

\usepackage{tikz}
\usetikzlibrary{shapes,decorations,arrows,calc,arrows.meta,fit,positioning}
\tikzset{
    -Latex,auto,node distance = 1 cm and 1 cm,semithick,
    state/.style ={ellipse, draw, minimum width = 0.7 cm},
    point/.style = {circle, draw, inner sep=0.04cm,fill,node contents={}},
    bidirected/.style={Latex-Latex,dashed},
    el/.style = {inner sep=2pt, align=left, sloped}
}
\definecolor{mygreen}{RGB}{0,153,0}
\definecolor{myred}{RGB}{153,0,0}

\usepackage{xurl}

\usepackage{enumitem}

\theoremstyle{definition}

\renewcommand\hl[1]{\ignorespaces#1}

\usepackage[noend]{algpseudocode}

\usepackage{subfiles}

\newcolumntype{P}[1]{>{\centering\arraybackslash}p{#1}}

\title{Estimating the Causal Effects of\\ Natural Logic Features in Transformer-Based NLI Models}

\name{Julia Rozanova$^{1}$, Marco Valentino$^{3}$, Andr\'e Freitas$^{1,2,3}$\\
$^{1}$Department of Computer Science, University of Manchester, UK\\  
$^{2}$ National Biomarker Centre, CRUK-MI, University of Manchester, UK\\
$^{3}$Idiap Research Institute, Switzerland \\ }

\address{}

\abstract{Rigorous evaluation of the causal effects of semantic features on language model predictions can be hard to achieve for natural language reasoning problems. However, this is such a desirable form of analysis from both an 
    interpretability and model evaluation perspective, that it is valuable
    to investigate specific patterns of reasoning with enough structure and regularity to identify and quantify systematic reasoning failures in widely-used models. In this vein, we pick a portion of the NLI task for which an explicit causal diagram can be systematically constructed: the case where across two sentences (the premise and hypothesis), two related words/terms occur in a shared context.
    In this work, we apply causal effect estimation strategies to measure the effect of \emph{context} interventions 
    (whose effect on the entailment label is mediated by the semantic \emph{monotonicity} characteristic) and interventions on the inserted 
    word-pair (whose effect on the entailment label is mediated by the \emph{relation} between these words).
    Extending related work on causal analysis of NLP models in different settings, we 
    perform an extensive interventional study on the NLI task to investigate  \emph{robustness to irrelevant changes} and \emph{sensitivity to impactful changes} of Transformers. The results strongly bolster the fact that similar benchmark accuracy scores may be observed for models that exhibit very different behaviour. Moreover, our methodology reinforces previously suspected biases from a causal perspective, including biases in favour of upward-monotone contexts and ignoring the effects of negation markers.}
    
\begin{document}

\maketitleabstract

    \section{Introduction}
    
    There is an abundance of reported cases where high accuracies in NLP tasks can be attributed to simple heuristics and dataset artifacts~\cite{mccoy}.
    As such, when we expect a language model to capture a specific reasoning strategy or correctly use certain semantic features, it has become good practice to perform evaluations that provide a more granular and qualitative view into model behaviour and efficacy.
    In particular, there is a trend in recent work to incorporate causal measures
    and \emph{interventional} experimental setups in order to better understand the
    captured features and reasoning mechanisms of NLP models ~\cite{vig-gender,finlayson-agreement,stolfo-maths,geiger-causal-abstraction,rozanova-etal-2023-interventional,arakelyan-etal-2024-semantic}.
   
    In general, it can be hard to pinpoint all the intermediate features and critical representation elements which are guiding the inference behind an NLP task. 
    However, in many cases there are subtasks which have enough semantic/logical regularity to perform stronger analyses and diagnose clear points of failure within larger tasks such as NLI and QA (Question Answering).
    As soon as we are able to draw a causal diagram which captures a portion of the model's expected reasoning capabillities, we may be guided in the design of interventional experiments which allow us to estimate 
    causal quantities of interest, giving insight into how 
    different aspects of the inputs are used by models.

In this work, we investigate a structured subset of the NLI task~\cite{rozanova-decomposing} to better understand the use of two semantic inference features by NLI models: concept
 relations and logical monotonicity.
We use these intermediate abstracted semantic feature labels to construct \emph{intervention sets} out of NLI examples which allow us to measure
certain causal effects. Building upon recent work on causal analysis of NLP models~\citet{stolfo-maths}, we use 
the intervention sets to systematically and quantitatively characterise models' \emph{sensitivity} to relevant changes in these semantic features and \emph{robustness} to irrelevant changes.

    Our contributions may be summarised as follows:
    \begin{itemize}
        \item Extending previous work on causal analysis of NLP models, we investigate a structured subproblem in NLI (in our case, a subtask based on natural logic~\cite{maccartney-manning}) and present a causal diagram which captures both desired and undesired potential reasoning routes which may describe model behaviour.
        \item We adapt the NLI-XY dataset of~\citet{rozanova-decomposing} to a meaningful collection of \emph{intervention sets} which enable the computation of certain causal effects.
        \item We calculate estimates for undesired direct causal effects and desired total causal effects, which also serve as a quantification of model robustness and sensitivity to our intermediate semantic features of interest.
        \item We compare a suite of BERT-like NLI models, identifying behavioural weaknesses in high-performing models and behavioural advantages in some worse-performing ones.
    \end{itemize}%
    
    To the best of our knowledge, we are the first to complement previous observations of models' brittleness with respect to context monotonicity with 
    the evidence of causal effect measures\footnote{Our code is available at \url{https://github.com/juliarozanova/counterfact_nli}.}, as  well as presenting new insights 
    that over-reliance on lexical relations is consequently also 
    tempered by the same improvement strategies.

    \section{Problem Formulation}
    
    \subsection{A Structured NLI Subtask} \label{sec:nli_subproblem}
    
    As soon as we have a concrete description of how a reasoning problem \emph{should} be treated, we can begin to evaluate how well a model emulates the expected behaviour and whether it is capturing the semantic abstractions at play. 
    
    In this work, we consider an NLI subtask which comes from the broader setting of \emph{Natural Logic} 
   ~\cite{maccartney-manning, humoss_polarity, sanchezvalencia}.
    As it has a rigid and well-understood structure, it is often used in interpretability and 
    explainability studies for NLI models~\cite{geiger-causal-abstraction, richardson_fragments, geiger-inducing, rozanova-decomposing, rozanova_supporting}. We begin with the format described in~\cite{rozanova-decomposing} 
    (we refer to this work for more detailed description and full definitions). 
    
    \noindent Consider two terms/concepts with a known \emph{relation label}, 
    such as one of the pairs: 
    
    \begin{table}[h!]
        \centering
        \begin{tabular}{ccc}
        \midrule
        Word/Term $x$ & Word/Term $y$  & Relation  \\
        \midrule
        brown sugar & sugar & $x \sqsubseteq y$ \\
        mammal & lion & $x \sqsupseteq y$ \\
        computer & pomegranate & $x \# y$ \\
        \midrule
    \end{tabular}
    \end{table}
    
    \noindent  Suppose the two terms occur in an identical context (comprising of a natural language sentence, like a template), for example:
    
    \begin{table}[h!]
        \centering
        \begin{tabular}{ll}
        \midrule
             Premise & I do not have any \textbf{sugar}.  \\
             Hypothesis & I do not have any \textbf{brown sugar}. \\\midrule
        \end{tabular}
    \end{table}
    
    A semantic property of the natural language context called \emph{monotonicity} determines whether there is an \emph{entailment 
    relation} between the sentences generated upon substitution/insertion
    of given related terms (formally, this is monotonicity in the sense 
    of preserving the ``order" between the inserted terms to an equally-directed entailment relation between the sentences.) 
    The context monotonicity may either be \emph{upward} ($\uparrow$) or
    \emph{downward} ($\downarrow$, as in the example above) or \emph{neither}. 
    
    The effect of the context's monotonicity in conjunction 
    with the relation between the inserted words 
    on the gold entailment label is summarised in table~\ref{table:value_summary}.
    \begin{table}[t!]
        \centering
        \resizebox{\columnwidth}{!}{
        \begin{tabular}{c|ccc}
        \backslashbox{$M$}{$R$} & $\sqsubseteq$ & $\sqsupseteq$ & $\#$ \\ \hline
             $\uparrow$ & Entailment & Non-Entailment &Non-Entailment \\ 
             $\downarrow$ &Non-Entailment& Entailment& Non-Entailment \\ 
        \end{tabular}%
        }
        \caption{The entailment gold labels as a function of two semantic features: the context montonicity (M) and the relation (R) of the inserted word pair.}
        \label{table:value_summary}
    \end{table}
    The authors of~\citet{rozanova-decomposing} provide a thus-formatted dataset called \emph{NLI-XY}, which we use as the basis for our causal effect estimation experiments.

Throughout the remainder of this paper, we will represent an NLI-$XY$ example $n$ as a tuple $n = (c, m, w, r, g)$ 
in which $c$ is the shared natural language context, $m$ is its monotonicity label,
$w$ is a pair ($w_1,w_2$) of nouns/noun phrases which will be inserted into the context (we refer to these as the \emph{inserted word pair} for brevity), $r$ is the concept inclusion relation label for $w$ and 
$g$ is the entailment gold label arising from $m$ and $r$ as per table {\ref{table:value_summary}}.
We denote by $P(Y \mid C=c, W=w)$ the probabilistic output of a trained NLI model
with the example $n$ as the NLI input (in particular, the input is the premise--hypothesis
pair ($c(w_1), c(w_2)$)). 

    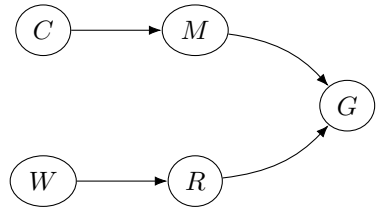
\begin{figure}
        \centering
        \begin{tabular}{cc}
             Variable & Description \\ \midrule
             $G$ & Gold Label\\
             $C$ &  Context \\
             $M$ &  Context Monotonicity\\
             $W$ &  Inserted Word Pair \\
             $R$ &  Word-Pair Relation \\
        \end{tabular}
       
        \vspace{1em}
        
        \begin{tikzpicture}[scale=2.0]
            \node[state] (c) at (0,2) {$C$};
            \node[state] (m) at (1,2) {$M$};
            
            \node[state] (w) at (0,1) {$W$};
            \node[state] (r) at (1,1) {$R$};
            \node[state] (g) at (2,1.5) {$G$};
    
            \path (c) edge (m);
            \path (m) edge[bend left=20] (g);
    
            \path (w) edge (r);
            \path (r) edge[bend right=20] (g);
            
        \end{tikzpicture}
        \caption{Causal Diagram for the Natural Logic Subtask}
        \label{fig:nli_xy_causal_diagram}
    \end{figure}

\hl{
As we have chosen a coarse segmentation of the monotonicity 
reasoning problem, we can present a simple causal diagram which illustrates 
our expectations for the correct reasoning scheme for a fixed 
class of NLI problems.}
\hl{The diagram in 
figure~{\ref{fig:nli_xy_causal_diagram}} shows the features on 
which the gold label is dependent on in the NLI-$XY$ dataset: only the context 
monotonicity $M$ and the concept pair relation $R$, 
which are respectively dependent on the content of the natural language context $C$ and the 
concept pair / word pair $W$ which is substituted into it. The exact values of the 
gold label with respect to these features may be referenced in 
table {\ref{table:value_summary}}.
}


    Naturally, it is always likely that models may fail to follow the described reasoning scheme for these NLI problems. In the next section (\ref{sec:model_causal_diagram}), we propose a causal diagram
    which also captures the reasoning possibilities an NLI model may follow, accounting for possible 
    confounding heuristics via unwanted direct effects.

\subsection{The Causal Structure of Model Decision-Making}\label{sec:model_causal_diagram}

In an ideal situation, a strong NLI model would identify the word-pair relation and the context 
monotonicity as the abstract variables relevant to the final entailment label.
In this case, these features would causally affect the model prediction in the same way 
they affect the gold label. 
Realistically, as shown in illuminating studies such as~\citet{mccoy-right-for-the-wrong},
models identify unexpected biases in the dataset and may end up using accidental 
correlations output labels, such as the frequency of certain words in a corpus.
For example,~\citet{mccoy-non-entailment} demonstrate how models can successfully exploit 
the presence of negation markers to anticipate non-entailment, even when it is not 
semantically relevant to the output label.

To ensure that the semantic features themselves are taken account into the model's output and not 
other surface-level confounding variables, one would like to perform interventional studies
which alter the value of the target feature but not other confounding variables.
This is, in many cases, not feasible (although attempts are sometimes made to at least perform 
interventions that only make minimal changes to the textual surface form, as in~\citet{kaushik-counterfactually-augmented}.)

~\citet{stolfo-maths} argue that it is useful to quantify instead the direct impact of irrelevant 
surface changes (controlling for values of semantic variables of interest) and compare them to 
\emph{total causal effects} of input-level changes: doing so, we may posit deductions about the 
flow of information via the semantic variables (or lack thereof).
For analyses where there is an attempt to align intermediate variables with explicit internals,
see \citet{vig-gender} and ~\citet{finlayson-agreement} for a mediation analysis approach, 
or~\citet{geiger-causal-abstraction} for an alignment strategy based on causal abstraction 
theory.

\paragraph{Diagram Specification}
We follow~\citet{stolfo-maths} in the strategy of explicitly modeling the
``irrelevant surface form" of the input text portions as variables in the causal diagram.
Their setting of \emph{math word problems} is decomposed into two compositional inputs: 
a question template and two integer arguments.
Our setting follows much the same structure: our natural language ``context'' plays the same role as 
their ``template'', but our arguments (an inserted word pair) have an additional layer of complexity as we also model the 
\emph{relation} between the arguments as an intermediate reasoning variable rather than 
the values themselves (as such, the structure of their template modeling in their causal diagram
is more applicable than the direct way they treat their numerical arguments.)

\begin{figure}[t]
\centering
\begin{tabular}{cc}
        Variable & Description \\ \midrule
        $Y$ & Model Prediction\\
        $G$ & Gold Label\\
        $C$ &  Context \\
        $M$ &  Context Monotonicity\\
        $S$ &  Context Textual Surface Form\\
        $W$ &  Inserted Word Pair \\
        $R$ &  Word-Pair Relation \\
        $T$ &  Word-Pair Textual Surface Form\\
\end{tabular}
\vspace{1em}

\begin{tikzpicture}[scale=1.5]
    \node[state] (c) at (0,2.5) {$C$};
    \node[state] (sc) at (1,3) {$S$};
    \node[state] (m) at (1,2) {$M$};
    
    \node[state] (w) at (0,0.5) {$W$};
    \node[state] (sw) at (1,0) {$T$};
    \node[state] (r) at (1,1) {$R$};
    \node[state] (g) at (2,1.5) {$G$};
    \node[state] (y) at (3,1.5) {$Y$};

    \path (c) edge (sc);
    \path (c) edge (m);
    \path[red] (sc) edge[bend left=20] (y);
    \path[mygreen] (m) edge[bend left=20] (g);
    \path[red] (m) edge[bend left=20] (y);

    \path (w) edge (sw);
    \path (w) edge (r);
    \path[red] (sw) edge[bend right=20] (y);
    \path[mygreen] (r) edge[bend right=20] (g);
    \path[red] (r) edge[bend right=20] (y);
    \path[mygreen] (g) edge (y);
    
\end{tikzpicture}
\caption{Specification of the causal diagram for possible routes of model reasoning for NLI-$XY$ problems. 
Green edges indicate \emph{desired} causal influence, while red edges indicate \emph{undesired} paths of causal influence via surface-level heuristics.} 
\label{fig:causal_diagram}
\end{figure}
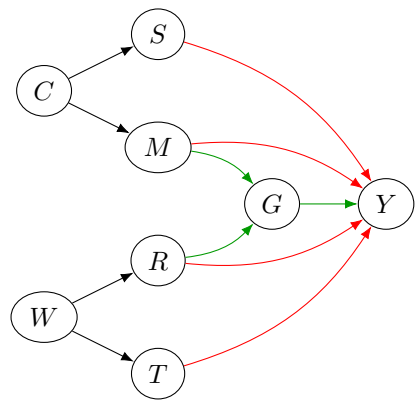

We present our own causal diagram in figure~\ref{fig:causal_diagram}. We introduce the textual context
$C$ as an input variable, which is further decomposed into more abstract variables: 
its \emph{monotonicity} $M$ (which directly affects the gold truth $G$) and the textual surface form \hl{$S$}
of the context .
The other input variable is the word-pair insertion which we will summarise as a single variable $W$. Once again, $W$ has a potential effect on the model decision through its textual surface form \hl{$T$} and 
\hl{via the} relation $R$ between the words. The gold truth $G$ is dependent on $M$ and $R$ only.  
Finally, the outcome variable is the model prediction $Y$.
The paths for which we would like to observe the highest causal effect are the paths to $Y$ from the inputs via $M, R$ and through the gold truth variable $G$.
However, each of $S, T, M $ and $R$ have direct links to the model output $Y$ as well 
(indicated in red):  these are potential direct effects which 
are \emph{unwanted}. 
\hl{For example, we would not want a model to learn a prediction heuristic
based directly on the variable $M$, such as consistently predicting non-entailment any time a 
downward monotone context is recognised. 
Similarly, a direct effect of $S$ or $T$ would look like a heuristic which predicts the entailment label purely based on the presence of words which 
happened to co-occur with that label in the training 
data.}
The key goal of this study is to compare the extent to which models exhibit the high 
causal effects for the \emph{desired} diagram routes and lower causal effects for the \emph{undesired} routes.

\section{Estimating the Causal Effects}\label{sec:causal_effects}

\hl{
Given a fixed set $N$ of NLI-$XY$ examples, we define an \emph{intervention} $\mathcal{I}$ on $N$ as a set of 
pairs $(n,n')$ of NLI-$XY$ examples for (one for each $n \in N$), where $n'=(c', m', w', r', g')$ is 
a second NLI-$XY$ example which represents a modified version of $n$ 
(in practice, a modification of either $c$ or $w$). We denote by $N'$ the set of modified NLI-$XY$ examples, 
so that $\mathcal{I} \subseteq N \times N'$.
}

\hl{For any pair $(n, n') \in \mathcal{I}$ , we define the 
change-of-prediction indicator
}

$$CP(n,n') =  
\begin{cases}
    1 &\text{if } y \not = y'\\
    0 &\text{if } y = y'\\
\end{cases}
,$$
\hl{where}
$$y = \arg \max_{i \in \{0,1\}} P(Y=i \mid C=c, W=w)$$
\hl{
(namely, the model prediction which assigns the entailment label with the highest predicted probability)
and}
$$y' = \arg \max_{i \in \{ 0, 1\}} P(Y=i \mid C=c',W=w').$$
\hl{
\mbox{\citet{stolfo-maths}} refer to the average change-of-prediction quantity for a given intervention 
$\mathcal{I}$ as a \emph{causal effect}. 
This causal effect quantity is named 
and interpreted differently depending on
the conditions of the intervention: in particular, which variables are changed and which are kept constant throughout the intervention set over which we will take the average.
}

\subsection{Interventions for Calculating TCE and DCE}
\hl{
The quantities of interest in~\mbox{\citet{stolfo-maths}} 
are the \emph{total causal effect} (TCE) of interventions on the variables 
which we would like to see having an effect on the prediction (in our case, $C$  and $W$)
and the \emph{direct causal effect} (DCE) of interventions on the variables which we do 
\emph{not} wish to unnecessarily impact the model prediction (in our case, $T$ and $S$).

For a given \emph{source} variable and \emph{target} variable, 
whether we are measuring a DCE or TCE differs only in the design of the intervention set, 
which in turn depends on the structure of the causal diagram.
For the design of the relevant intervention sets, we follow the strategy 
in~\mbox{\citet{stolfo-maths}}, as the upper portion of their causal diagram 
(concerning the natural language question template, its textual surface form and the implicit math operation) is equivalent to both the upper and lower half of our diagram in figure~{\ref{fig:causal_diagram}}.

In this work, we provide four intervention sets: $\mathcal{I}_0, \mathcal{I}_1, 
\mathcal{I}_2, \mathcal{I}_3$, each corresponding to the quantities 
$\mbox{TCE } (C ~\mbox{on } Y), \mbox{TCE } (W ~\mbox{on } Y), \mbox{DCE } (T \to Y)$ and $ \mbox{DCE } (S \to Y)$ respectively.
\footnote{\hl{To be consistent with the notation in ~\mbox{\citet{stolfo-maths}}, we will stylize these quantities as (for example) TCE($C ~\mbox{on } Y$)
and DCE($S \to Y$), where the arrow emphasizes that the quantity is specific to a direct path in the causal diagram (passing through no intermediate variables).}}
We stick to their nomenclature of total causal effect (TCE) and direct causal effect 
(DCE), but define the quantities in the way that they are concretely calculated
(in both our experiments and in~\mbox{\citet{stolfo-maths}}): as an \emph{estimate} 
of the causal effect quantity, which they present as an expected value 
of the change-of-prediction indicator. 
}


\paragraph{(Desired) Total Causal Effects}
\hl{We estimate the total causal effect of the context $C$ on the model prediction $Y$ 
by constructing an intervention set $\mathcal{I}_0$ as follows:
starting with a randomly sampled set $N$ of NLI-$XY$ examples, 
we intervene on each $n \in N$ by sampling a 
different context $c'$ from the NLI-$XY$ dataset which should 
result in a changed prediction, while keeping 
the inserted word pair $w$ constant. In summary, every $(n,n') \in \mathcal{I}_0$ satisfies }
$$(c\not = c', m \not= m', w = w', r = r', g \not = g').$$
\hl{We then calculate:}
$$~\mbox{TCE}  (C ~\mbox{on } Y) = \frac{1}{\lvert \mathcal{I}_0 \rvert } \sum_{(n,n') \in I_0} CP(n,n') $$ 

\hl{
Secondly, we estimate the total causal effect of the inserted word pair $W$ on the model prediction
$Y$ by constructing an intervention set $\mathcal{I}_1$ as follows:
starting with a randomly sampled set $N$ of NLI-$XY$ examples, 
we intervene on each $n \in N$ by sampling a 
different inserted word pair $w'$ from the NLI-$XY$ dataset which should 
result in a changed prediction, while keeping 
the shared context $c$ constant.  In summary, every $(n,n') \in \mathcal{I}_1$ satisfies 
}
$$(c = c', m = m', w \not = w', r \not = r', g \not = g').$$
\hl{We then calculate:}
$$~\mbox{TCE}  (W ~\mbox{on } Y) = \frac{1}{\lvert \mathcal{I}_1 \rvert } \sum_{(n,n') \in I_1} CP(n,n') $$ 

\hl{
Following ~\mbox{\citet{stolfo-maths}}, we interpret this quantity as a measure of 
model \emph{sensitivity} to relevant context (respectively, inserted word pair) changes.
As it quantifies how often the prediction changes when it should, we would like to see this value being as close to $1$ as possible.
}

\paragraph{(Undesired) Direct Causal Effects}

\hl{The total causal effect does not distinguish whether this effect is mediated through the 
preferred causal route (for example, via context's monotonicity) or through a model heuristic 
based on the textual surface form: it is taking into account all possible routes of influence.
}
The key suggestion in~\citet{stolfo-maths} is that even though we have no 
feasible intervention strategies which allow us to calculate the causal effect of the 
intermediate variables $M$ and $R$ on $Y$ as mediated through the gold label $G$ 
(the effect of greatest interest to us), 
we may yield some insight into their causal influence by comparing the 
relevant TCE to the unwanted \emph{direct causal effect} DCE ($S \to Y$) (
respectively, DCE ($T \to Y$)).

\hl{
To estimate the direct causal effect of the textual surface form $S$ of the context $C$ which is irrelevant to the context monotonicity $M$, 
we construct an intervention set $\mathcal{I}_2$ as follows:
starting with a randomly sampled set $N$ of NLI-$XY$ examples, we intervene
on each $n \in N$ by sampling a different context $c'$ from the NLI-$XY$ dataset
while conditioning on the monotonicity (specifically, $c'$ is chosen so that
its monotonicity attribute $m'$ is the same as that of $c$). The word pair 
$w'$ (and therefore its relation $r'$) are kept the same as in $n$, so the prediction is expected \emph{not} to change.
In summary, every $(n, n') \in \mathcal{I}_{2}$ satisfies
}
$$(c \not = c', m  = m', w = w', r = r', g = g').$$

\hl{We then calculate:}
$$~\mbox{DCE}  (S ~\to Y) = \frac{1}{\lvert \mathcal{I}_2 \rvert } \sum_{(n,n') \in \mathcal{I}_2} CP(n,n') $$ 
\hl{To estimate the direct causal effect of the textual surface form $T$ of the 
inserted word pair $W$ which is irrelevant to the word pair relation $R$, 
we construct an intervention set $\mathcal{I}_3$ as follows:
starting with a randomly sampled set $N$ of NLI-$XY$ examples, we intervene
on each $n \in N$ by sampling a different inserted word pair $w'$ from the 
NLI-$XY$ dataset while conditioning on the word pair relation 
(specifically, $w'$ is chosen so that
its relation attribute $r'$ is the same as that of $w$). 
The context $c'$ (and therefore its monotonicity $m'$) are kept the same as in $n$, so the prediction is expected \emph{not} to change.
In summary, every $(n, n') \in \mathcal{I}_{3}$ satisfies
}
$$(c = c', m = m', w \not = w', r = r', g = g').$$
\hl{We then calculate:}
$$~\mbox{DCE}  (T ~\to Y) = \frac{1}{\lvert \mathcal{I}_3 \rvert } \sum_{(n,n') \in \mathcal{I}_3} CP(n,n') $$ 

\hl{
Once again following~\mbox{\citet{stolfo-maths}}, we interpret this quantity as a measure of model \emph{robustness}
to irrelevant context (respectively, inserted word pair) changes.
As it quantifies how often the prediction changes in cases when it \emph{shouldn't},
we would like to see this value being as close to $0$ as possible.
} We present examples and dataset statistics for the intervention sets in the next section, 
along with the summary of the intervention schema in table \ref{table:intervention_schema}.

\section{Experimental Setup}
\begin{table*}[htb!]
\resizebox{\textwidth}{!}{
\begin{tabular}{@{}p{1.7cm}lp{1.7cm}p{5cm}p{5cm}lll@{}}
\toprule
Intervention Set & Target Quantity                     & Intervention Step & Premise                       & Hypothesis                            & M            & R             & G              \\ \midrule
\multirow{2}{*}{$\mathcal{I}_1$} & \multirow{2}{*}{TCE($W ~\mbox{on } Y$)}     & Before       & There's a cat on the pc.   & There's a cat on the machine.      & $\uparrow$   & $\sqsubseteq$ & Entailment     \\
                                && After        & There's a cat on the tree. & There's a cat on the fruit tree.     & $\uparrow$   & $\sqsupseteq$ & Non-Entailment \\ \midrule
\multirow{2}{*}{$\mathcal{I}_3$} &\multirow{2}{*}{DCE($T \to Y$)} & Before       & There are no students yet.    & There are no first-year students yet. & $\downarrow$ & $\sqsupseteq$ & Entailment     \\
                                && After        & There are no people yet.     & There are no women yet.                & $\downarrow$ & $\sqsupseteq$ & Entailment     \\ \midrule
\end{tabular}%
}
\caption{Example word-pair insertion interventions for determining the total causal effect of label-relevant word-pair changes and the direct causal effect of
label-irrelevant word-pair changes.}
\label{table:insertion_int_examples}
\end{table*}

\begin{table*}[htb!]
\resizebox{\textwidth}{!}{
\begin{tabular}{@{}p{1.7cm}lp{1.7cm}p{5cm}p{5cm}lll@{}}
\toprule
Intervention Set & Target Quantity                     & Intervention Step & Premise                       & Hypothesis                            & M            & R             & G              \\ \midrule
\multirow{2}{*}{$\mathcal{I}_0$} &\multirow{2}{*}{TCE($C ~\mbox{on } Y$)}     & Before       & You can't live without fruit .                       & You can't live without strawberries .                & $\uparrow$   & $\sqsupseteq$ & Non-Entailment \\

                                && After        & All fruit study English.                              & All strawberries study \hl{English}.                       & $\downarrow$ & $\sqsupseteq$ & Entailment     \\ \midrule
\multirow{2}{*}{$\mathcal{I}_2$} &\multirow{2}{*}{DCE($S \to Y$)} & Before       & He has no interest in seafood .                        & He has no interest in oysters .                        & $\downarrow$ & $\sqsupseteq$ & Entailment     \\
                                && After        & I don't want to argue about this in front of seafood . & I don't want to argue about this in front of oysters . & $\downarrow$ & $\sqsupseteq$ & Entailment     \\ \bottomrule
\end{tabular}%
}
\caption{Example context interventions for determining the total causal effect of label-relevant context changes and the direct causal effect of
label-irrelevant context changes.}
\label{table:context_int_examples}
\end{table*}

\subsection{Data and Interventions}\label{sec:intervention_scheme}


\begin{table}[htb!]
\centering
\resizebox{0.9\columnwidth}{!}{
\begin{tabular}{@{}p{1.7cm}lllllllp{2.5cm}@{}}
\toprule
Intervention Set & Target Measure& $C$ & $W$ & $M$ & $R$ & $G$ & & Interventions in Dataset\\ \midrule
\hl{$\mathcal{I}_0$} &TCE $(C \to Y)$ &$\not=$ & $=$ & $\not =$ & $=$ & $\not=$ & & 14270\\
\hl{$\mathcal{I}_1$} &TCE $(W \to Y)$ & $=$ & $\not=$ & $=$ & $\not =$ & $\not =$ & &  22640\\
\bottomrule
\hl{$\mathcal{I}_2$} &DCE $(S \to Y)$ & $\not =$ & $=$ & $=$ &$=$ & $=$ & & 20910\\
\hl{$\mathcal{I}_3$} &DCE $(T \to Y)$ & $=$ & $\not =$ & $=$ & $=$ & $=$ & & 25960\\
\midrule
\end{tabular}%
}
\caption{Intervention schema and dataset statistics: which variables are held constant and which are changed
in the construction of intervention sets for the calculation of the indicated effects.}
\label{table:intervention_schema}
\end{table}

We use the NLI-$XY$ evaluation dataset to construct intervention pairs $(n,n')$ by using a sampling/filtering strategy 
as in~\cite{stolfo-maths} according to the 
intervention schema in table~\ref{table:intervention_schema}.
In particular, for constructing \emph{context} interventions, we sample a seed set of 400 NLI-$XY$
premise/hypothesis pairs. This is the \emph{pre-intervention} NLI example. 
For each, we fix the insertion pair and filter through the NLI-$XY$ dataset for all 
examples with the shared insertion pair but different context, conditioned as necessary 
on the properties of the other variables as in the intervention schema. For insertion pairs, we do the opposite.
The number of interventions we produce in this way for our experiments are reflected in the last column of 
table~\ref{table:intervention_schema}. In summary, the changes are context replacements and related word-pair replacements; 
we provide text-level examples in tables~\ref{table:insertion_int_examples} and~\ref{table:context_int_examples} .

\subsection{Model Choice and Benchmark Comparison}\label{sec:models}

We include the following models \footnote{All pretrained models are from the Huggingface \emph{transformers} library \cite{transformers},
except for infobert and the pretrained model counterparts fine-tuned on HELP: their sources are linked in the README of the accompanying code.} in our study: firstly, the models evaluated in NLI-$XY$ paper~\cite{rozanova-decomposing}, namely
\hl{roberta-large-mnli, facebook/bart-large-mnli, bert-base-uncased-snli} and their counterparts fine-tuned on the HELP dataset~\cite{yanaka-help}
Next, the infobert model, which is trained on three benchmark training
sets of interest: MNLI~\cite{mnli}, SNLI~\cite{snli} and ANLI~\cite{anli} (currently at the top of the leaderboard for the adversarial ANLI test set, as of January 2023)
Lastly, another roberta-large checkpoint, also trained on all three benchmark NLI training training sets (as well as FEVER-NLI~\cite{fever}).
We report their scores on the mentioned benchmark datasets alongside 
the relevant total and direct causal effects we are interested in.

Note that as the HELP dataset is a \hl{two-class} entailment dataset (as opposed to datasets like MNLI, which 
are \hl{three-class}), we cannot directly compare existing reported scores. 
As such, we adapt the \hl{three-class} scores to a \hl{two-class} score by grouping two of the \hl{three-class} labels 
(``contradiction" and ``neutral") into the \hl{two-class} umbrella label "non-entailment".


\section{Results and Discussion}
We examine and compare the results for the models listed in~\ref{sec:models}.
We first observe the word-pair insertion intervention experiments in~\ref{sec:insertion_results}, then the context intervention experiments in~\ref{sec:context_results} and finally present a categorical overview of these results in section~\ref{sec:comparison_results}, contextualised by benchmark scores.

\subsection{Causal Effect of Inserted Word Pairs} \label{sec:insertion_results}
\begin{figure}[t!]
\centering
\includegraphics[width=\columnwidth]{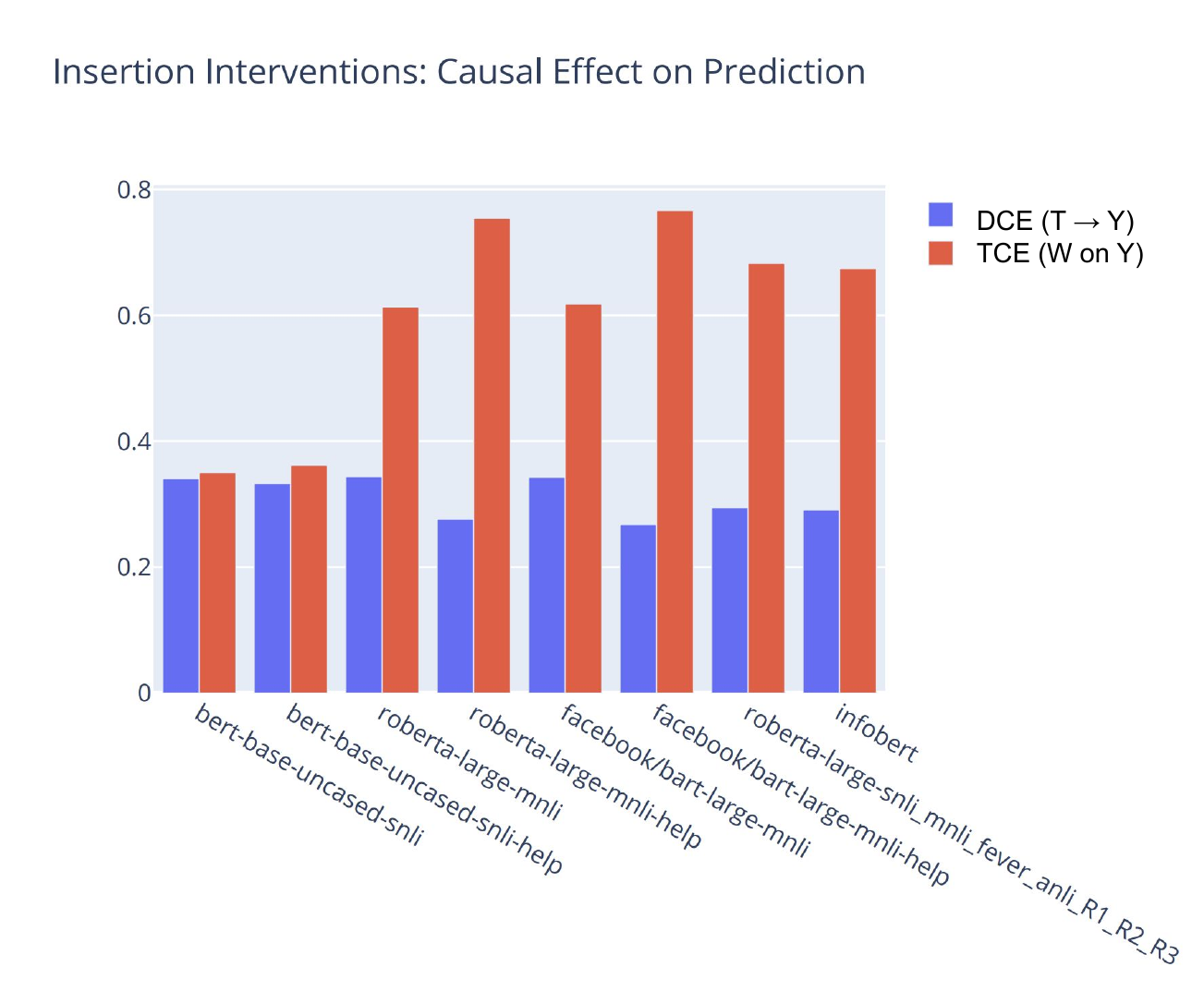}
\resizebox{\columnwidth}{!}{
\begin{tabular}{@{}p{5.5cm}lrrlrr@{}}
\toprule
Model                                             &  & \multicolumn{1}{l}{DCE($T \to Y$)} & \multicolumn{1}{l}{TCE($W ~\mbox{on}~  Y$)} &  & \multicolumn{1}{l}{TCE/DCE Ratio} & \multicolumn{1}{l}{Delta} \\ \midrule
bert-base-uncased-snli                            &  & 0.341                                             & 0.350                                          &  & 1.027                          & 0.009                  \\
bert-base-uncased-snli-help                       &  & 0.332                                             & 0.361                                          &  & 1.087                           & 0.029                  \\
roberta-large-mnli                                &  & 0.343                                             & 0.613                                          &  & 1.785                          & 0.269                  \\
roberta-large-mnli-help                           &  & 0.276                                             & 0.754                                          &  & 2.730                          & 0.478                  \\
facebook/bart-large-mnli                          &  & 0.342                                             & 0.618                                          &  & 1.805                          & 0.275                  \\
facebook/bart-large-mnli-help                     &  & 0.268                                             & 0.766                                          &  & 2.863                          & 0.499                  \\
roberta-large-snli\_mnli\_fever\_anli\_R1\_R2\_R3 &  & 0.294                                             & 0.682                                            &  & 2.321                          & 0.388                  \\
infobert                                          &  & 0.291                                             & 0.674                                          &  & 2.320                          & 0.384                  \\ \bottomrule
\end{tabular}%
}
\caption{Results for Insertion Interventions}
\label{fig:insertion_experiment}
\end{figure}

The results for the substituted word-pair intervention experiment are reported in
figure~\ref{fig:insertion_experiment}.
The most desireable outcome is a DCE($T \to Y$) which is \emph{as low as possible}
in combination with a TCE($W ~\mbox{on}~Y$) which is \emph{as high as possible}.
The lower this DCE, the higher the model robustness to \emph{irrelevant word pair surface form} changes. On the other hand, the higher the specified TCE, the greater
the model's sensitivity to \emph{word pair insertion changes affecting the gold label}. 

The largest delta between these two quantities can be seen in the 
roberta-large-mnli-help and facebook-bart-large-mnli-help models.
This is important to note: the HELP dataset~\cite{yanaka-help} is explicitly designed 
to bolster model success on natural logic problems, but until now there has been 
little to no evidence that it improves the treatment of word-pair relations.
In particular, the internal probing results in~\citet{rozanova-decomposing} show that
probing performance for the intermediate word-pair relation label decreases slightly
for roberta-large-mnli after fine-tuning on HELP; as such, it was thought
that the HELP improvements on natural logic could solely be attributed to improved 
context monotonicity treatment.
Now, however, we observe distinct improvements in robustness to irrelevant word-pair 
insertion changes and sensitivity to relevant ones.

More generally, the work in~\citet{rozanova-decomposing} does indicate that the large MNLI-based 
models are already very successful in distinguishing the relation between substituted words.
The word-pair relation label has a high \emph{probing} result for all of these models, as well as strong signs of systematicity in their error analysis. 
This is in line with our observations of relatively large deltas between the DCE and TCE here, compared to the smaller BERT-based models.  

\subsection{Causal Effect of Contexts}\label{sec:context_results}

\begin{figure}[t!]
\centering
\includegraphics[width=\columnwidth]{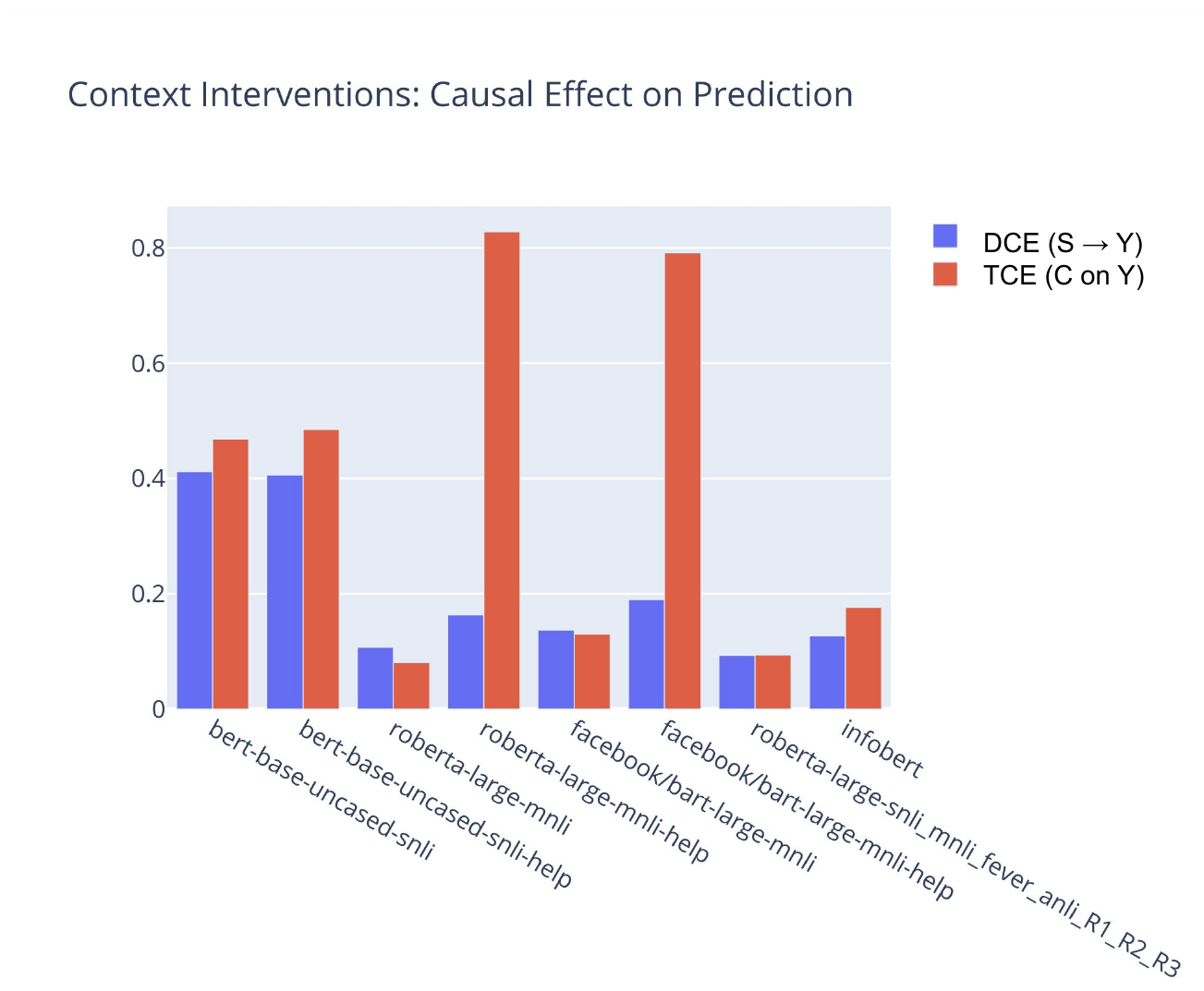}
\resizebox{\columnwidth}{!}{
\begin{tabular}{@{}p{5.5cm}lrrlrr@{}}
\toprule
Model                                             &  & \multicolumn{1}{l}{DCE($S \to Y$)} & \multicolumn{1}{l}{TCE($C ~\mbox{on}~ Y$)} &  & \multicolumn{1}{l}{TCE/DCE Ratio} & \multicolumn{1}{l}{Delta} \\ \midrule
bert-base-uncased-snli                            &  & 0.412                             & 0.468                           &  & 1.136                           & 0.0563                  \\
bert-base-uncased-snli-help                       &  & 0.406                             & 0.485                            &  & 1.194                           & 0.079                  \\
roberta-large-mnli                                &  & 0.107                             & 0.081                           &  & 0.751                          & -0.027                 \\
roberta-large-mnli-help                           &  & 0.163                             & 0.828                           &  & 5.070                          & 0.665                  \\
facebook/bart-large-mnli                          &  & 0.136                              & 0.130                           &  & 0.954                          & -0.006                 \\
facebook/bart-large-mnli-help                     &  & 0.189                             & 0.791                           &  & 4.167                          & 0.601                  \\
roberta-large-snli\_mnli\_fever\_anli\_R1\_R2\_R3 &  & 0.093                             & 0.093                           &  & 1.008                          & 0.001                   \\
infobert                                          &  & 0.127                             & 0.176                           &  & 1.385                           & 0.049                  \\ \bottomrule
\end{tabular}%
}
\caption{Results for Context Interventions}
\label{fig:context_experiment}
\end{figure}
The results for the context intervention experiments are reported in
figure~\ref{fig:context_experiment}.
The most desireable outcome is a DCE($S \to Y$) which is \emph{as low as possible}
in combination with a TCE($C ~\mbox{on}~ Y$) which is \emph{as high as possible}.
For context interventions, we start to see major distinctions in the sensitivity of models to 
important context changes - especially the effect of the HELP fine-tuning 
dataset in increasing model reasoning with respect to context structure.
In line with previous behavioural findings in~\citet{richardson_fragments, yanaka-help, yanaka-med, geiger-partially, rozanova-decomposing} and all the way back to~\citet{glue}, which observe systematic
failure of large language models in downward monotone contexts, we notice that all
of the models trained only on the large benchmarks sets fail to correctly change their prediction when a context change requires it to do so 
(as indicated by the low TCE score). In~\citet{yanaka-help},~\citet{rozanova-decomposing} and~\citet{rozanova_supporting},  the positive effect of the HELP dataset is already evident, 
but here we may also compare it to roberta-large-mnli tuned on many additional training sets, precluding the possibility that its 
helpfulness can be attributed only to a larger amount of training data.   

We note that although the situation of the TCE/DCE ratio for 
roberta-large-mnli being less than one may seem peculiar, it is important to 
keep in mind that the intervention sets used for estimating these quantities are sampled independently so some margin of error is warranted.
As in~\citet{stolfo-maths}, we interpret this result to simply mean that the 
causal influence is comparable whether we are affecting the ground truth result (as in the TCE($C ~\mbox{on}~  Y$) case) or not (as in the DCE($S \to Y$) case).


\subsection{Benchmark Scores and Causal Effects}\label{sec:comparison_results}

\begin{table*}[]
\resizebox{\textwidth}{!}{
\begin{tabular}{@{}lllllllllllllll@{}}
\toprule
\multicolumn{1}{c}{\multirow{2}{*}{\textbf{Model}}} &  & \multicolumn{6}{l}{NLI Benchmark Evaluation (2 Class Accuracy)}                                     &  & \multicolumn{2}{l}{Context Changes} &  & \multicolumn{2}{l}{Inserted Word-Pair Changes} &  \\ \cmidrule(lr){2-8} \cmidrule(lr){10-11} \cmidrule(lr){13-14}
\multicolumn{1}{c}{}                                &  & SNLI           & MNLI-M         & MNLI-MM        & ANLI-R1        & ANLI-R2        & ANLI-R3        &  & Robustness       & Sensitivity      &  & Robustness             & Sensitivity           &  \\ \midrule
bert-base-uncased-snli                              &  & 0.766          & 0.620          & 0.623          & 0.567          & 0.596          & 0.580          &  & Mid              & Mid              &  & Mid                    & Low                   &  \\
bert-base-uncased-snli-help                         &  & 0.757          & 0.627          & 0.626          & 0.505          & 0.508          & 0.546          &  & Mid              & Mid              &  & Mid                    & Low                   &  \\
facebook/bart-large-mnli                            &  & 0.935          & 0.940          & 0.939          & 0.596          & 0.563          & 0.593          &  & High             & Low              &  & Mid                    & Mid                   &  \\
facebook/bart-large-mnli-help                       &  & 0.727          & 0.802          & 0.795          & 0.538          & 0.489          & 0.528          &  & Mid/High         & \textbf{Highest} &  & \textbf{Highest}       & \textbf{Highest}      &  \\
roberta-large-mnli                                  &  & 0.931          & 0.941          & 0.940          & 0.614          & 0.529          & 0.5325         &  & \textbf{Highest} & Lowest           &  & Mid                    & Mid                   &  \\
roberta-large-mnli-help                             &  & 0.738          & 0.668          & 0.656          & 0.565          & 0.554          & 0.574          &  & High             & \textbf{Highest} &  & \textbf{Highest}       & \textbf{Highest}      &  \\
roberta-large-snli\_mnli\_fever\_anli   &  & 0.949          & 0.936          & 0.939          & 0.810          & 0.659          & 0.666          &  & \textbf{Highest} & Lowest           &  & Mid                    & Mid/High              &  \\
infobert                                            &  & \textbf{0.950} & \textbf{0.943} & \textbf{0.941} & \textbf{0.837} & \textbf{0.682} & \textbf{0.683} &  & High             & Low              &  & Mid                    & Mid/High              &  \\ \bottomrule
\end{tabular}%
}
\caption{Overall 2 class accuracy on original NLI benchmarks and qualitative comparison against the performed causal intervention analysis. The accuracy is not necessarily predictive of the performances achieved using a systematic causal inspection.}
\label{table:comparison_table}
\end{table*}

A summary of the performance of all models on popular benchmarks
alongside a categorical breakdown of robustness and sensitivity is presented in table~\ref{table:comparison_table}. 
The robustness/sensitivity categories are a qualitative assessment, identifying the \emph{lowest} and \emph{highest} scores within a category, and categorising other models correspondingly as \emph{low}, \emph{mid} or \emph{high} performers for the given categories. The sensitivity property is tied to the desired total causal effect, while the robustness property is tied to the undesired direct causal effect (note in particular that the latter is judged as \emph{inversely proportional:} the model with the lowest given DCE is judged the ``highest" in terms of robustness).

The key observation is that the models which achieve the highest 
performance on benchmarks may be far from the best performers with 
respect to our quantitative markers of strong reliance of important
causal features.
In particular, models such as \emph{infobert} are outperformed in 
our behavioural causal effect analyses by weaker models that are
fine-tuned on a relatively small helper dataset such as \emph{HELP}. 
It is important to note that such changes coincide with drops in benchmarks performance too, 
but any model interventions that discourage the exploitation of heuristics (evident from a lower DCE for surface form features) may have that effect. 
    

\section{Related Work}
\paragraph{Natural Logic Handling in NLI Models}
It has been known for some time that large NLI models are frequently 
tripped up by downward-monotone reasoning~\cite{richardson-fragments, glue, yanaka-help, rozanova-decomposing, geiger-partially}.
Various datasets have been created to evaluate and improve this behaviour, such as 
HELP~\cite{yanaka-help}, MoNLI~\cite{geiger-partially}, MQNLI~\cite{geiger_posing_fair}, MED~\cite{yanaka-med}. 
~\citet{rozanova-decomposing} introduced NLI-XY, secondary compositional dataset built from portions of MED, where the intermediate features of \emph{context monotonicity} and \emph{concept relations} are explicitly 
labelled: this is the dataset we use in this work. 
Non-causal structural analyses of model internals with respect to natural 
logic features include~\citet{rozanova-decomposing} (a probing study), but we leave to the next section some existing works where natural logic intersects 
with the world of causal approaches to NLP. 

\paragraph{Causal Analysis in NLP}
Causal modelling has appeared in NLP works in various forms, such as the 
investigations of the causal influence of data statistics~\cite{elazar_causal_data} and 
mediation analyses~\cite{vig-gender, finlayson-agreement} which link intermediate linguistic/semantic features to model internals.
~\citet{stolfo-maths}, our core reference, appears to be the first to 
use explicitly causal effect measures as indicators of 
sensitivity and robustness (for some non-causal approaches to measuring model robustness in NLP, we point to~\citet{textfooler} and~\citet{checklist}).
For a fuller summary of the use of causality in NLP, please see the 
survey by~\citet{feder-causal-nlp}.
Specific to natural logic, works with causal approaches include~\citet{geiger-partially} (which perform interchange interventions at a token representation level),~\citet{geiger-causal-abstraction} (where an ambitious causal abstraction experiment attempts to align model internals with candidate causal models) and the works of~\citet{geiger-partially} and~\citet{wu-proxy}, (where attempts are made to build a prescribed causal structure into models themselves). In particular,~\citet{wu-proxy} create a  ``causal proxy model" which becomes the basis for a new explainable predictor designed to replace the original neural network. 

\section{Conclusion}

The results strongly bolster the fact that similar benchmark accuracy scores may be observed for models that exhibit very different behaviour, 
especially concerning specific semantic reasoning patterns and 
higher-level properties such as robustness/sensitivity to target features.
In this work, we have been able to causally investigate previously suspected biases    
in NLI models.
For example, previous observations~\cite{rozanova-decomposing, yanaka-med} that roberta-large-mnli is biased in favour of assuming upward-monotone contexts, ignoring the effects of things like negation markers, agrees with our observations that it exhibits poor context sensitivity. 
Furthermore, the causal flavour of the study adds a complimentary narrative to works that investigate model internals via probing~\cite{rozanova-decomposing} 
and observe the presence/absence of intermediate semantic features in the models' representation.
Instead of merely suggesting that these features are captured, we can gain insight into their causal influence via connected causal effect estimates. The causal measures presented here show us that even the highest-performing models can
systematically fail to adapt their predictions to changing context structure, suggesting an over-reliance on word relations across the premise and hypothesis.
Finally, we have also added the observation that existing strategies to improve responsiveness to context changes also increase the \emph{robustness} word-pair insertion changes.

\section*{Acknowledgements}
This work was partially funded by the Swiss National Science Foundation (SNSF) project NeuMath (\href{https://data.snf.ch/grants/grant/204617}{200021\_204617}), by the EPSRC grant EP/T026995/1 entitled “EnnCore: End-to-End Conceptual Guarding of Neural Architectures” under Security for all in an AI enabled society, by the CRUK National Biomarker Centre, and supported by the Manchester Experimental Cancer Medicine Centre.

\bibliographystyle{lrec-coling2024-natbib}
\bibliography{refs2}

\end{document}